\documentclass[11pt,a4paper]{article}
\usepackage{acl2015}
\usepackage{times}
\usepackage{url}
\usepackage{latexsym}
\usepackage{amsmath,bm}
\usepackage{amsfonts}
\usepackage{mathrsfs}
\usepackage{amsbsy}
\usepackage{graphicx}
\usepackage{xcolor}
\usepackage{multirow}
\usepackage{paralist}
\usepackage{subfigure}

\makeatletter
\newcommand{\verbatimfont}[1]{\def\verbatim@font{#1}}%
\makeatother

\title{Classifying Relations via Long Short Term Memory Networks\\
along Shortest Dependency Paths}
\date{}

\author{Yan Xu,$\!\!^\dag$ Lili Mou,$\!\!^\dag$ Ge Li,$\!^\dag$\thanks{\ \ Corresponding authors.}
\ Yunchuan Chen,$\!\!^\ddagger$ Hao Peng,$\!\!^\dag$ Zhi Jin$^{\dag}{}^*$\\
$^\dag$Software Institute, Peking University, 100871, P. R. China\\
\{xuyan14,lige,zhijin\}@sei.pku.edu.cn,\{doublepower.mou,penghao.pku\}@gmail.com\\
$^\ddagger$University of Chinese Academy of Sciences, chenyunchuan11@mails.ucas.ac.cn
}
\date{}

\begin{document}
\maketitle

\begin{abstract}
Relation classification is an important research arena in
the field of natural language processing (NLP).
In this paper, we present SDP-LSTM, a novel neural network to classify the relation
of two entities in a sentence.
Our neural architecture leverages the shortest dependency path (SDP) between two entities;
multichannel recurrent neural networks, with long short term memory (LSTM) units,
pick up heterogeneous information along the SDP.
Our proposed model has several distinct features: (1)
The shortest dependency paths retain
most relevant information (to relation classification),
while eliminating irrelevant words in the sentence. (2)
The multichannel LSTM networks allow
effective information integration from heterogeneous sources over the dependency paths.
(3) A customized dropout strategy regularizes the neural network to alleviate overfitting.
We test our model on the SemEval 2010 relation classification task,
and achieve an $F_1$-score of 83.7\%, higher than competing methods in the literature.

\end{abstract}

\section{Introduction}\label{sIntroduction}

Relation classification is an important NLP task.
It plays a key role in various scenarios,
e.g., information extraction \cite{openIE}, question answering \cite{QAFreebase},
medical informatics \cite{MedicalRE}, ontology learning \cite{ontology}, etc.
The aim of relation classification is to categorize into predefined classes the relations between pairs of marked entities in given texts.
For instance, in the sentence ``A trillion gallons of
[water]$_{e_1}$ have been poured into an empty [region]$_{e_2}$ of outer space,'' the entities \textit{water} and
\textit{region} are of relation \verb|Entity-Destination|$({e_1}, {e_2})$.

Traditional relation classification approaches rely largely on feature representation
\cite{MaxEntRE}, or kernel design \cite{KerRE,SpdKernel}.
The former method usually incorporates a large set of features;
it is difficult to improve the model performance
if the feature set is not very well chosen.
The latter approach, on the other hand,
depends largely on the designed kernel, which
summarizes all data information.
Deep neural networks, emerging recently,
provide a way of highly automatic feature learning \cite{RL},
and have exhibited considerable potential \cite{CNN,RankCNN}.
However, human engineering---that is, incorporating human knowledge to the network's architecture---is still
important and beneficial.

\begin{figure}[!t]
\centering
\includegraphics[width=.45\textwidth]{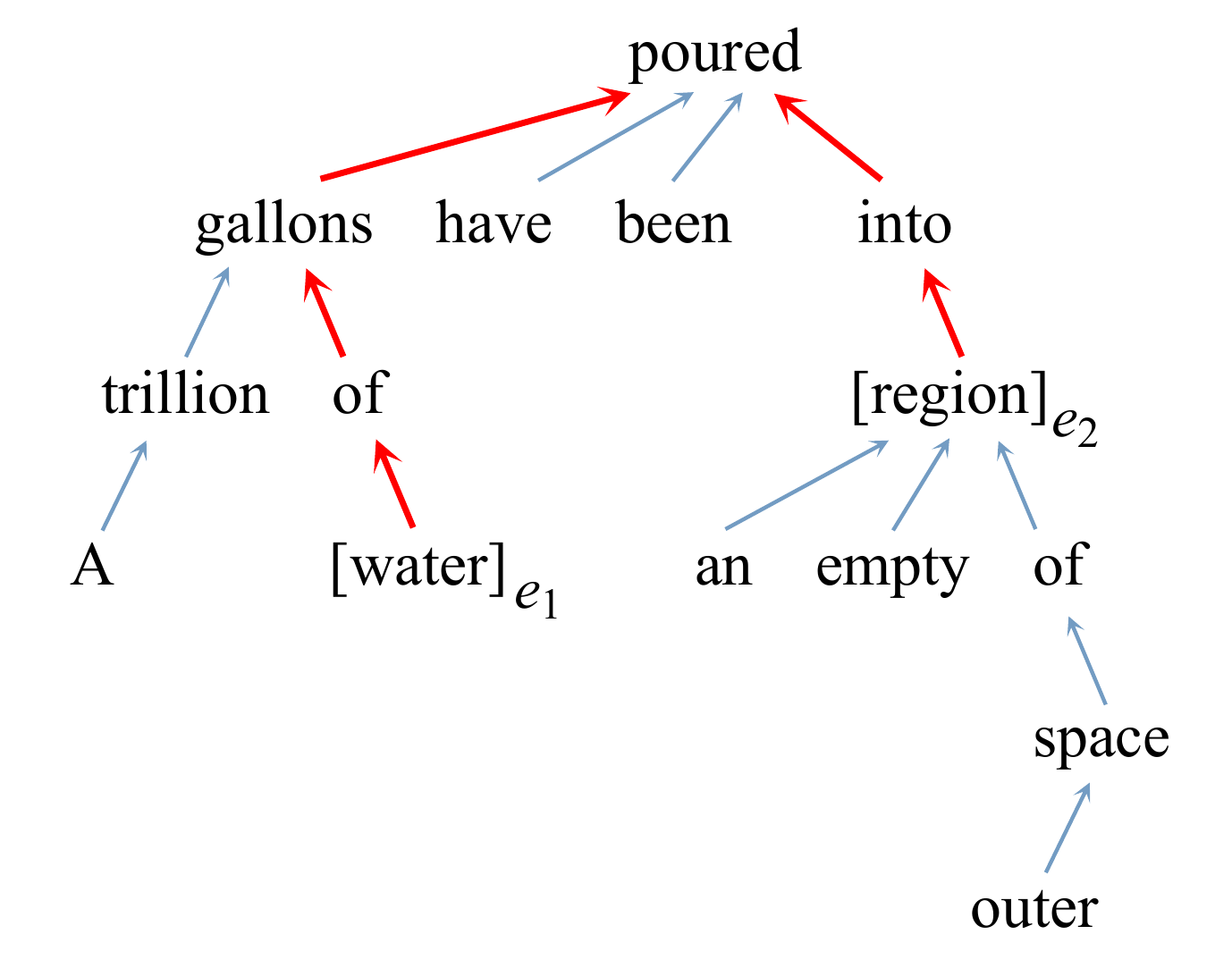}
\caption{The dependency parse tree corresponding to the sentence
``A trillion gallons of water have been poured
into an empty region of outer space.'' Red lines indicate the shortest dependency path
 between entities \textit{water} and \textit{region}. 
 An edge $a\rightarrow b$ refers to $a$ being governed by $b$. 
 Dependency types are labeled by the parser, but not presented in the figure for clarity.}\label{fDependency}
\end{figure}

This paper proposes a new neural network, SDP-LSTM, for relation classification.
Our model utilizes the shortest dependency path (SDP) between two entities in a sentence;
we also design a long short term memory (LSTM)-based recurrent neural network for
information processing.
The neural architecture is mainly inspired by the following observations.
\begin{itemize}
\item Shortest dependency paths are informative \cite{relex,deriving}. To determine the two entities' relation,
we find it mostly sufficient
to use only the words along the SDP: they concentrate on most relevant information while diminishing less relevant noise.
Figure \ref{fDependency} depicts the dependency parse tree of the aforementioned sentence.
Words along the SDP form a trimmed phrase (\textit{gallons of water poured into region}) of the original sentence,
which conveys much information about the target relation.
Other words, such as \textit{a}, \textit{trillion}, \textit{outer space},
are less informative and may bring noise if not dealt with properly.

\item Direction matters. Dependency trees are a kind of directed graph. The dependency relation between \textit{into} and \textit{region} is \verb|PREP|; such relation hardly makes any sense if the directed edge is reversed.
Moreover, the entities' relation distinguishes its directionality, that is,
$r(a,b)$ differs from $r(b,a)$, for a same given relation $r$ and two entities $a, b$.
Therefore, we think it necessary to let the neural model
process information in a direction-sensitive manner.
Out of this consideration, we separate an SDP into two sub-paths,
each from an entity to the common ancestor node.
The extracted features along the two sub-paths are concatenated
to make final classification.

\item Linguistic information helps.
For example, with prior knowledge of hyponymy, we know ``water is a kind of substance.''
This is a hint that the entities, \textit{water} and \textit{region},  are more of
\verb|Entity-Destination| relation than, say, \verb|Communication-Topic|.
To gather heterogeneous information along SDP, we design a multichannel recurrent neural network.
It makes use of information from various sources, including words themselves, POS tags, WordNet hypernyms,
and the grammatical relations between governing words
and their children. 
\end{itemize}

For effective information propagation and integration,
our model leverages LSTM units during recurrent propagation.
We also customize a new dropout strategy
for our SDP-LSTM network to alleviate the problem of overfitting.
To the best of our knowledge, we are the first to use
LSTM-based recurrent neural networks for the relation classification task.

We evaluate our proposed method on the SemEval
2010 relation classification task,
and achieve an $F_1$-score of 83.7\%, higher than
competing methods in the literature.

In the rest of this paper, we 
review related work in Section \ref{sRelatedwork}.
In Section \ref{sModel}, we describe our SDP-LSTM model in detail.
Section \ref{sExperiment} presents quantitative experimental results.
Finally, we have our conclusion in Section \ref{sConclusion}.


\section{Related Work}\label{sRelatedwork}

Relation classification is a widely studied task in the NLP community.
Various existing methods mainly fall into three classes:
feature-based, kernel-based, and neural network-based.

In feature-based approaches,
different sets of features are extracted
and fed to a chosen classifier (e.g., logistic regression).
Generally, three types of features are often used.
Lexical features concentrate on the entities of interest,
e.g., entities \textit{per se}, entity POS,
entity neighboring information.
Syntactic features include chunking, parse trees, etc.
Semantic features are exemplified by
the concept hierarchy, entity class, entity mention.
\newcite{MaxEntRE} uses a maximum entropy model
to combine these features for relation classification.
However, different sets of handcrafted features are largely complementary to each other
(e.g., hypernyms versus named-entity tags),
and thus it is hard to improve performance in this way \cite{REknowledge}.

Kernel-based approaches specify some measure of similarity between two data samples, without explicit feature representation.
\newcite {KerRE} compute the similarity of two trees by utilizing their common subtrees.
\newcite {SpdKernel} propose a shortest path dependency kernel for relation classification. Its main idea is that the relation strongly relies on the dependency path between two given entities. \newcite{reexamine} provides a systematic analysis of several kernels 
and show that relation extraction can benefit from combining convolution kernel and syntactic features.
\newcite {EmbedTreeK} introduce semantic information into kernel methods 
in addition to considering structural information only.
One potential difficulty of kernel methods is that
all data information is completely summarized by the kernel function (similarity measure),
and thus designing an effective kernel becomes crucial.

Deep neural networks, emerging recently, can learn underlying features automatically,
and have attracted growing interest in the literature.
\newcite{RAE} propose a recursive neural network (RNN) along sentences' parse trees
for sentiment analysis;
such model can also be used to classify relations \cite{MVRNN}.
\newcite{CustomRNN} explicitly weight phrases' importance in RNNs to improve performance.
\newcite{chainRNN} rebuild an RNN on the dependency path between two marked entities.
\newcite{CNN} explore convolutional neural networks,
by which they utilize sequential information of sentences.
\newcite{RankCNN} also use the convolutional network; besides, they propose a ranking loss function with data cleaning, and achieve the state-of-the-art result in SemEval-2010 Task 8.

In addition to the above studies,
which mainly focus on relation classification approaches and models,
other related research trends include information extraction from Web documents in a semi-supervised manner
\cite{MiniSupRE,OIE-1}, dealing with small datasets without enough labels
by distant supervision techniques
\cite{DisSupRE}, etc.

\section{The Proposed SDP-LSTM Model}\label{sModel}
\begin{figure*}
\ \includegraphics[width=.97\textwidth]{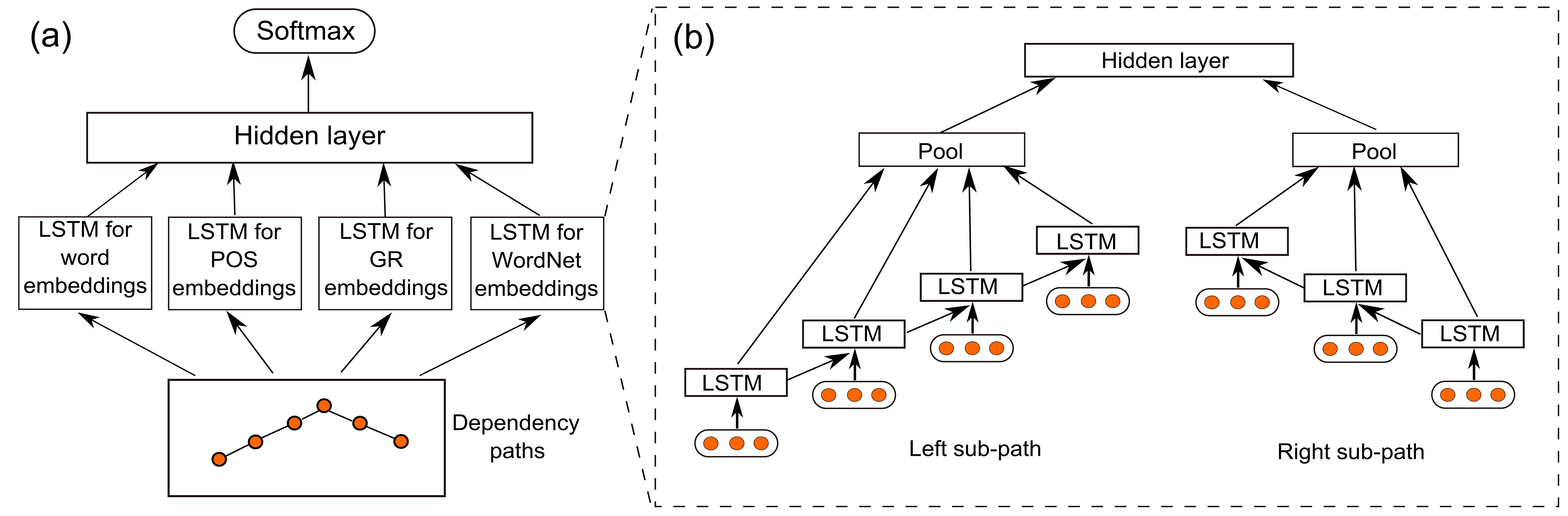}
\caption{(a) The overall architecture of SDP-LSTM.
(b) One channel of the recurrent neural networks built upon the shortest dependency path.
The channels are words, part-of-speech (POS) tags, grammatical relations
(abbreviated as \textit{GR} in the figure),
and WordNet hypernyms.
}\label{fArchitecture}
\end{figure*}

In this section, we describe our SDP-LSTM model in detail.
Subsection \ref{ssOverview} delineates the overall architecture
of our model.
Subsection \ref{ssSDP} presents the rationale of using SDPs.
Four different information channels along
the SDP are explained in Subsection \ref{ssChannels}.
Subsection \ref{ssLSTM} introduces the
recurrent neural network with long short term memory, which
is built upon the dependency path. Subsection \ref{ssDropout} customizes
a dropout strategy for our network to alleviate overfitting.
We finally present our training objective in Subsection \ref{ssObjective}.

\subsection{Overview}\label{ssOverview}

Figure \ref{fArchitecture}
depicts the overall architecture of our SDP-LSTM network.

First, a sentence is parsed to a dependency tree by the Stanford parser;\footnote{
http://nlp.stanford.edu/software/lex-parser.shtml} the shortest dependency path (SDP) is extracted as the input of our network.
Along the SDP, four different types of
information---referred to as \textit{channels}---are used, including the words, POS tags,
grammatical relations, and WordNet hypernyms.
(See Figure \ref{fArchitecture}a.)
In each channel, discrete inputs, e.g., words,
are mapped to real-valued vectors, called
\textit{embeddings}, which capture the underlying
meanings of the inputs.

Two recurrent neural networks (Figure \ref{fArchitecture}b)
pick up information along the left and right
sub-paths of the SDP, respecitvely. (The path is separated by the common ancestor node of two entities.)
Long short term memory (LSTM) units
are used in the recurrent networks
for effective information propagation.
A max pooling layer thereafter gathers information
from LSTM nodes in each path.

The pooling layers from different channels are concatenated,
and then connected to a hidden layer.
Finally, we have a $\operatorname{softmax}$ output layer for classification.
(See again Figure \ref{fArchitecture}a.)

\subsection{The Shortest Dependency Path}\label{ssSDP}

The dependency parse tree is naturally
suitable for relation classification because
it focuses on the action and agents in a sentence \cite{ImageSent}.
Moreover, the shortest path between entities, as discussed in Section \ref{sIntroduction},  condenses most illuminating information for entities' relation.

We also observe that the sub-paths, separated by
the common ancestor node of two entities,
provide strong hints for the relation's directionality.
Take Figure \ref{fDependency} as an example.
Two entities \textit{water} and \textit{region}
have their common ancestor node, \textit{poured}, which
separates the SDP into two parts:
$$\text{[water]}_{e_1}\rightarrow\text{of}\rightarrow\text{gallons}\rightarrow
\text{poured}$$
and
$$\text{poured}\leftarrow\text{into}\leftarrow\text{[region]}_{e_2}$$
The first sub-path captures information of $e_1$, whereas the second sub-path
is mainly about $e_2$.
By examining the two sub-paths separately, we know $e_1$ and $e_2$ are
of relation \verb|Entity-|\verb|Destination|$(e_1, e_2)$, rather
than \verb|Entity-Destination|$(e_2, e_1)$.

Following the above intuition,
we design two recurrent neural networks,
which propagate bottom-up from the entities to their common ancestor.
In this way, our model is direction-sensitive.

\subsection{Channels}\label{ssChannels}
We make use of four types of information along the SDP for relation classification.
We call them \textit{channels} as these information sources do not interact during
recurrent propagation. 
Detailed channel descriptions are as follows.

\begin{itemize}
\item \textbf{Word representations}. Each word in a given sentence is mapped to a real-valued vector by looking up 
in a word embedding table. Unsupervisedly trained on a large corpus, word embeddings are thought to be able to well capture words' syntactic and semantic information \cite{2013Mikolov}.
\item \textbf{Part-of-speech tags}. Since word embeddings are obtained on
a generic corpus of a large scale, the information they contain may not agree with a specific sentence. We deal with this problem by allying each input word with its POS tag,
 e.g., \verb|noun|, \verb|verb|, etc. In our experiment, we only take into use a coarse-grained POS category, containing 15 different tags.

\item \textbf{Grammatical relations}.
The dependency relations between a governing word and its children
makes a difference in meaning. 
A same word pair may have different dependency relation types.
For example,
``\textit{beats} $\xrightarrow{\operatorname{nsubj}}$ \textit{it}''
 is distinct from
``\textit{beats} $\xrightarrow{\operatorname{dobj}}$ \textit{it}.''
Thus, it is necessary to capture such grammatical relations in SDPs.
In our experiment, grammatical relations are grouped into 19 classes, mainly based on a coarse-grained 
classification \cite{TypeDep}.
\item \textbf{WordNet hypernyms}. As illustrated in Section \ref{sIntroduction},
hyponymy information is also useful for relation classification.
(Details are not repeated here.)
To leverage WordNet hypernyms,
we use a tool developed by
\newcite{SeqTagger}.\footnote{http://sourceforge.net/projects/supersensetag}
The tool assigns a hypernym to each word, from 41 predefined concepts in WordNet,
e.g., \verb|noun.food|, \verb|verb.motion|, etc.
Given its hypernym, each word gains a more abstract concept, which helps to build a linkage between different but conceptual similar words.
\end{itemize}

As we can see, POS tags, grammatical relations, and WordNet hypernyms
are also discrete (like words \textit{per se}).
However, no prevailing embedding learning method exists
for POS tags, say.
Hence, we randomly initialize their embeddings, and tune them in a supervised fashion
during training. We notice that these information sources contain much fewer
symbols, 15, 19, and 41, than the vocabulary size (greater than 25,000). Hence, we believe
our strategy of random initialization is feasible, because they can be
adequately tuned during supervised training.

\subsection{Recurrent Neural Network with Long Short Term Memory Units}\label{ssLSTM}

The recurrent neural network is suitable for modeling sequential data
by nature, as it keeps a hidden state vector $\bm h$, which changes with input
data at each step accordingly. We use the recurrent network to gather information
along each sub-path in the SDP (Figure \ref{fArchitecture}b).

The hidden state $\bm h_t$, for the $t$-th word in the sub-path, is a function of its previous
state $\bm h_{t-1}$ and the current word $\bm x_t$.
Traditional recurrent networks 
have a basic interaction, that is, the input is linearly transformed by a weight matrix and non-linearly squashed by an activation function.
Formally, we have
$$\bm h_t=f(W_{in}\bm x_t+W_{rec}\bm h_{t-1}+\bm b_h)$$
where $W_{in}$ and $W_{rec}$ are weight matrices for the input and
recurrent connections, respectively.
$\bm b_h$ is a bias term for the hidden state vector,
and $f_h$ a non-linear activation function (e.g., $\operatorname{tanh}$).

One problem of the above model is known as
\textit{gradient vanishing or exploding}.
The training of neural networks requires gradient back-propagation.
If the propagation sequence (path) is too long,
the gradient may probably either grow, or decay, exponentially,
depending on the magnitude of $W_{rec}$.
This leads to the difficulty of training.

\begin{figure}[!t]
\includegraphics[width=.45\textwidth]{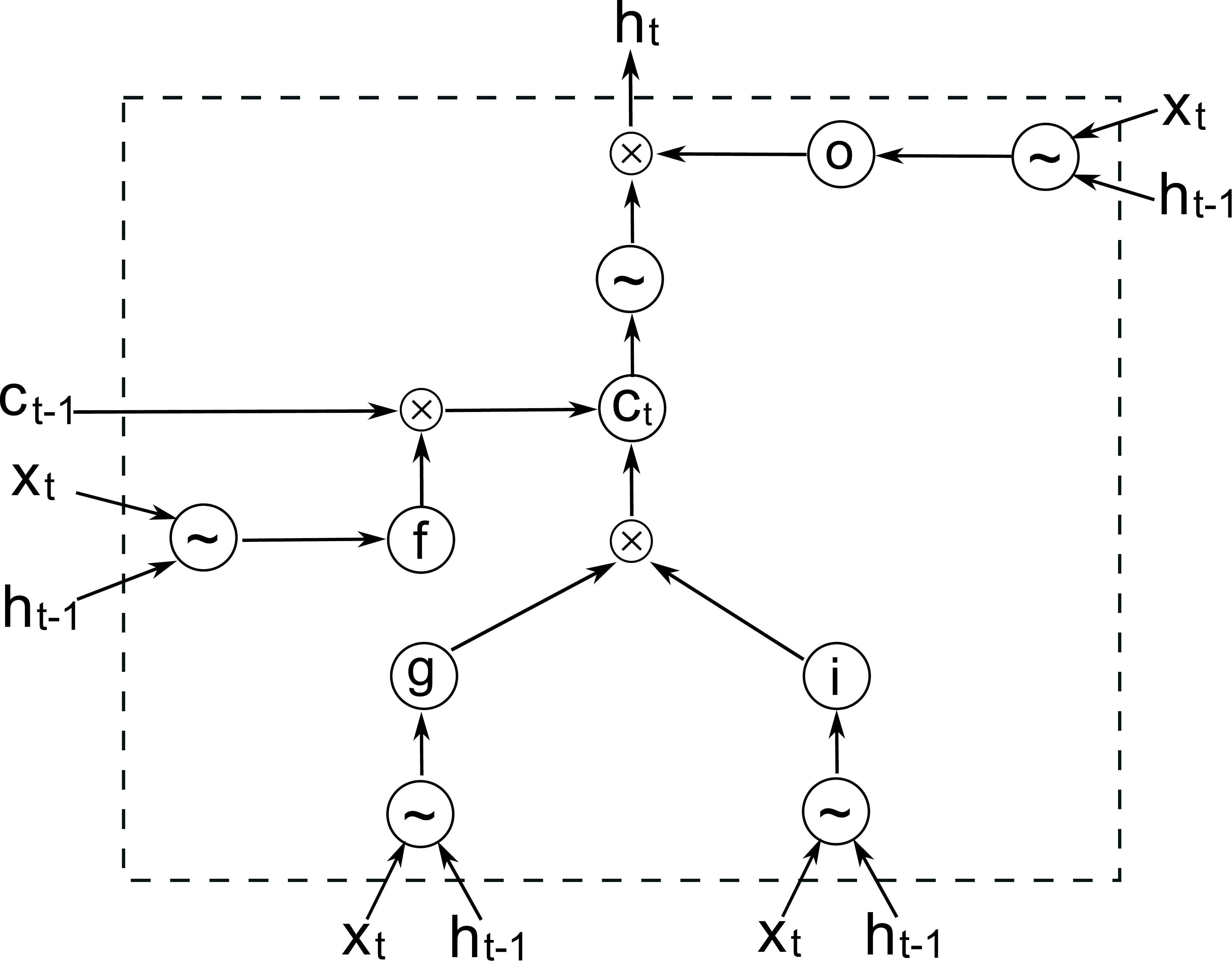}
\caption{A long short term memory unit.
$h$: hidden unit. $c$: memory cell. $i$: input gate.
$f$: forget gate. $o$: output gate.
$g$: candidate cell. $\otimes$: element-wise multiplication.
$\sim$: activation function.}\label{fLSTM}
\end{figure}

Long short term memory (LSTM) units are proposed in \newcite{1998LSTM}
to overcome this problem. The main idea is to introduce an adaptive
gating mechanism,
which decides the degree to which LSTM units keep the previous state and
memorize the extracted features of the current data input.
Many LSTM variants have been proposed in the literature.
We adopt in our method a variant introduced by \newcite{2014LearnToExecute},
also used in \newcite{TreeLSTM}.

Concretely, the LSTM-based recurrent neural network
comprises four components: an input gate ${\bm i_t}$, a forget gate ${\bm f_t}$, an output gate ${\bm o_t}$, and a memory cell ${\bm c_t}$ (depicted in Figure \ref{fLSTM} and formalized through
Equations \ref{elstm1}--\ref{elstm6} as bellow).

The three adaptive gates $\bm i_t$, $\bm f_t$, and $\bm o_t$
depend on the previous state $\bm h_{t-1}$ and the current input $\bm x_t$
(Equations \ref{elstm1}--\ref{elstm3}).
An extracted feature vector $\bm g_t$ is also computed, by Equation \ref{elstm4},
serving as the candidate memory cell.
\begin{align}
\bm i_t &= \sigma(W_i\!\cdot\!\bm x_t+U_i\!\cdot\!\bm h_{t-1}+\bm b_i)        \label{elstm1}\\
\bm f_t &= \sigma(W_f\!\cdot\!\bm x_t+U_f\!\cdot\!\bm h_{t-1}+\bm b_f)        \label{elstm2}\\
\bm o_t &= \sigma(W_o\!\cdot\!\bm x_t+U_o\!\cdot\!\bm h_{t-1}+\bm b_o)        \label{elstm3}\\
\bm g_t &= \tanh(W_g\cdot\!\bm x_t+U_g\!\cdot\!\bm h_{t-1}+\bm  b_g)     \label{elstm4}
\end{align}


The current memory cell $\bm c_t$ is a  combination of
the previous cell content $\bm c_{t-1}$ and the candidate content $\bm g_{t}$,
weighted by the input gate $\bm i_t$ and forget gate $\bm f_t$, respectively.
(See Equation \ref{elstm5} below.)
\begin{equation}
\bm c_t =\bm i_t\otimes\bm g_t+\bm f_t\otimes\bm c_{t-1}    \label{elstm5}
\end{equation}

The output of LSTM units is the the recurrent network's hidden state, which
is computed by Equation \ref{elstm6} as follows.
\begin{equation}
\bm h_t =\bm o_t\otimes\tanh(\bm c_t)\label{elstm6}
\end{equation}

In the above equations,  $\sigma$ denotes a $\operatorname{sigmoid}$ function;
$\otimes$ denotes element-wise multiplication.

\subsection{Dropout Strategies}\label{ssDropout}

A good regularization approach is needed to alleviate overfitting. Dropout, proposed recently by \newcite{DropoutFirst}, 
has been very successful on feed-forward networks. By randomly omitting feature detectors from the network during training, it can obtain less interdependent network units and achieve better performance. However, the conventional dropout does not work well with recurrent neural networks with LSTM units, since dropout may hurt the valuable memorization ability of memory units.

As there is no consensus on how to drop out LSTM units in the literature,
we try several dropout strategies for our SDP-LSTM network:
\begin{itemize}
\item Dropout embeddings;
\item Dropout inner cells in memory units, including $i_t$, $g_t$, $o_t$, $c_t$, and $h_t$; and
\item Dropout the penultimate layer.
\end{itemize}
As we shall see in Section \ref{ssSetting},
dropping out LSTM units turns out to be inimical to our model, whereas
the other two strategies boost in performance.

The following equations formalize the dropout operations
on the embedding layers, where ${D}$ denotes the dropout operator.
Each dimension in the embedding vector, $\bm x_t$, is set to zero with a predefined dropout rate.
\begin{align}
\bm i_t &= \sigma(W_i\!\cdot\!D(\bm x_t)+U_i\!\cdot\!\bm h_{t-1}+\bm b_i)\label{edrop1}\\
\bm f_t &= \sigma(W_f\!\cdot\!D(\bm x_t)+U_f\!\cdot\!\bm h_{t-1}+\bm b_f)\label{edrop2}\\
\bm o_t &= \sigma(W_o\!\cdot\!D(\bm x_t)+U_o\!\cdot\!\bm h_{t-1}+\bm b_o)\\
\bm g_t &= \tanh\!\Big(W_g\!\cdot\!D(\bm x_t)+U_g\!\cdot\!\bm h_{t-1}+\bm b_g\Big)
\end{align}

\subsection{Training Objective}\label{ssObjective}

The SDP-LSTM described above
propagates information along a sub-path from an entity
to the common ancestor node (of the two entities).
A max pooling layer packs, for each sub-path, the recurrent network's states, $\bm h$'s,
to a fixed vector by taking the maximum value in each dimension.

Such architecture applies to all channels, namely, words, POS tags, grammatical relations,
and WordNet hypernyms. The pooling vectors in these channels
are concatenated, and fed to a fully connected hidden layer.
Finally, we add a $\operatorname{softmax}$ output layer for classification.
The training objective is the penalized cross-entropy error, given by
$$J=-\sum_{i=1}^{n_c} t_i\log y_i+\lambda\!\left(\sum_{i=1}^\omega\|W_i\|_F^2+
\sum_{i=1}^{\upsilon}\|U_i\|_F^2\!\right)$$
where $\bm t\in\mathbb{R}^{n_c}$ is the one-hot represented ground truth and
$\bm y\in\mathbb{R}^{n_c}$
is the estimated probability for each class by $\operatorname{softmax}$.
(${n_c}$ is the number of target classes.)
$\|\cdot\|_F$ denotes the Frobenius norm of a matrix; $\omega$ and $\upsilon$
are the numbers of weight matrices (for $W$'s and $U$'s, respectively).
$\lambda$ is a hyperparameter that specifies the magnitude of penalty on weights.
Note that we do not add $\ell_2$ penalty to biase parameters.

We pretrained word embeddings by \verb|word2vec| \cite{Word2vce}
on the English Wikipedia corpus;
other parameters are initialized randomly.
We apply stochastic gradient descent (with mini-batch 10)
for optimization; gradients are computed by standard back-propagation.
Training details are further introduced in Section \ref{ssSetting}.

\section{Experiments}\label{sExperiment}

In this section, we present our experiments in detail.
Our implementation is built upon \newcite{tbcnn}.
Section \ref{ssData} introduces the dataset;
Section \ref{ssSetting} describes hyperparameter settings.
In Section \ref{ssResult}, we compare
SDP-LSTM's performance with other methods in the literature.
We also analyze the effect of different channels in Section \ref{ssChannel}.

\subsection{Dataset}\label{ssData}

The SemEval-2010 Task 8 dataset is a widely used benchmark for
relation classification \cite{2010SVM}. 
The dataset contains 8,000 sentences for training, and 2,717 for testing.
We split 1/10 samples out of the training set for validation.

The target contains 19 labels: 9 directed relations, and an undirected \verb|Other| class.
The directed relations are list as below.

\smallskip
\begin{compactitem}
\item \verb|Cause-Effect|

\item \verb|Component-Whole|

\item \verb|Content-Container|

\item \verb|Entity-Destination|

\item \verb|Entity-Origin|

\item \verb|Message-Topic|

\item \verb|Member-Collection|

\item \verb|Instrument-Agency|

\item \verb|Product-Producer|
\end{compactitem}

\smallskip
In the following are illustrated two sample sentences with directed relations.

\begin{quote}
[People]$_{e_1}$ have been moving back into [downtown]$_{e_2}$.

Financial [stress]$_{e_1}$ is one of the main causes of [divorce]$_{e_2}$.

\end{quote}
The target labels are \verb|Entity-Destination| ($e_1, e_2$),
and \verb|Cause-Effect|($e_1, e_2$), respectively.

The dataset also contains an undirected \verb|Other| class.
Hence, there are 19 target labels in total.
The undirected \verb|Other| class takes in entities that do not
fit into the above categories, illustrated by the following example.

\begin{quote}
A misty [ridge]$_{e_1}$ uprises from the [surge]$_{e_2}$.
\end{quote}

We use the official macro-averaged $F_1$-score to evaluate model performance.
This official measurement excludes the \verb|Other| relation.
Nonetheless, we have no special treatment of \verb|Other| class in our experiments,
which is typical in other studies.


\begin{figure*}[!t]
\subfigure[Dropout word embeddings\!\!\!\!\!\!]
{\includegraphics[width=.33\textwidth]{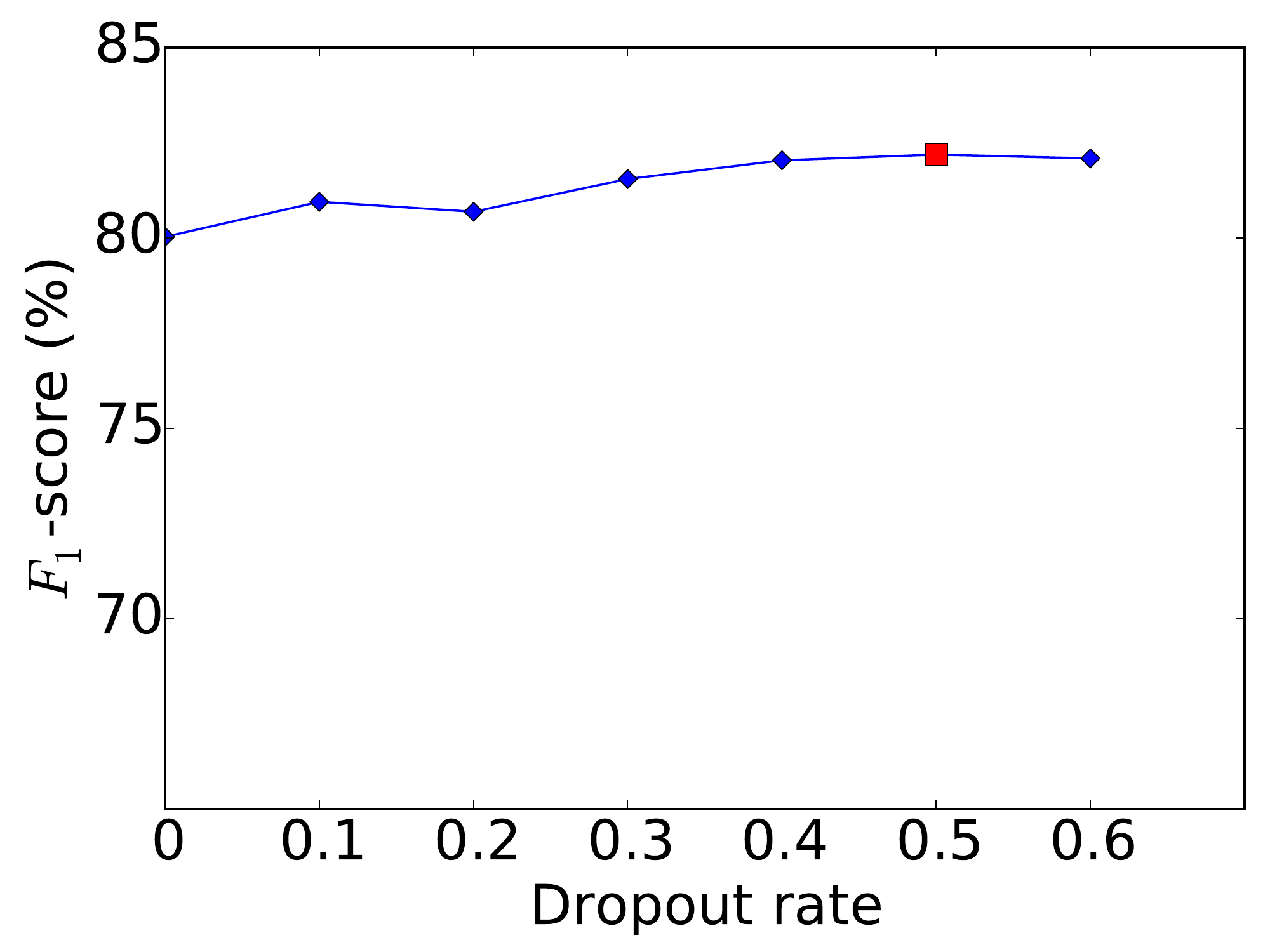}}
\subfigure[Dropout inner cells of memory units\!\!\!\!\!\!\!]{\includegraphics[width=.33\textwidth]{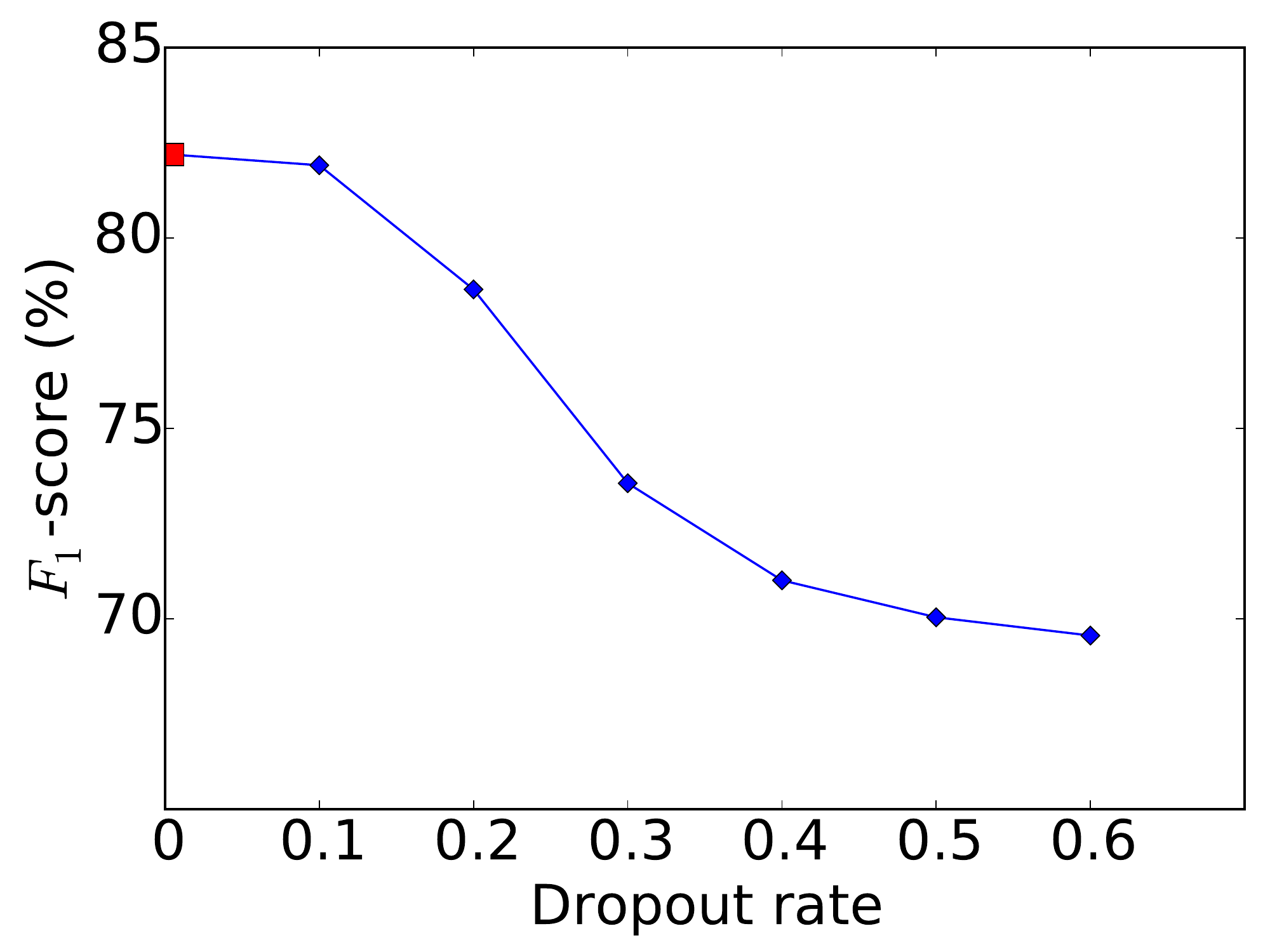}}
\subfigure[Dropout the penultimate layer\!\!\!\!\!\!\!\!\!]{\includegraphics[width=.33\textwidth]{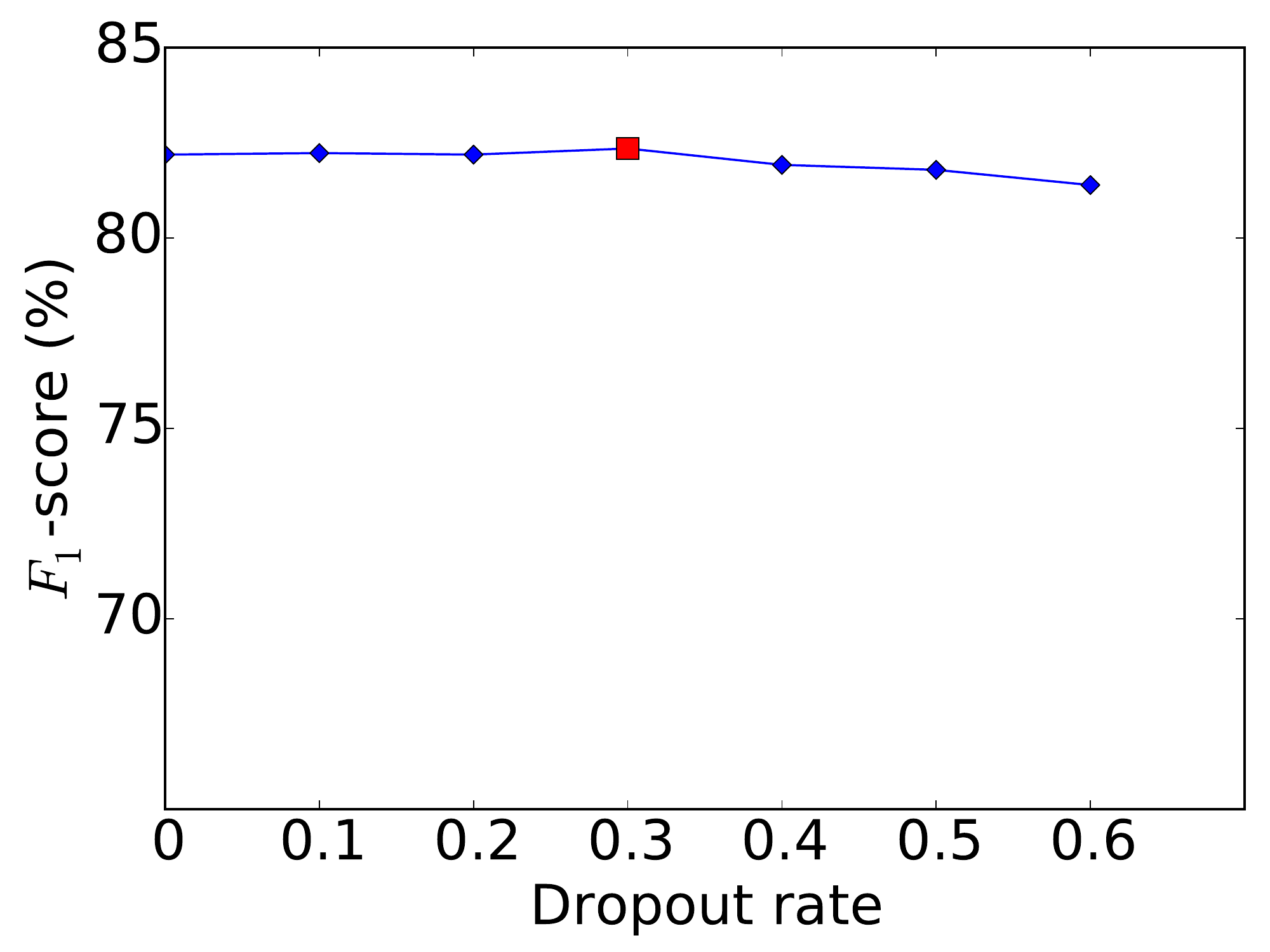}}
\caption{$F_1$-scores versus dropout rates. 
We first evaluate the effect of dropout embeddings (a). 
Then the dropout of the inner cells (b) and the penultimate layer (c) is tested with
word embeddings being dropped out by 0.5.}
\label{fDropout}
\end{figure*}

\subsection{Hyperparameters and Training Details}\label{ssSetting}
This subsection presents hyperparameter tuning for our model.
We set word-embeddings to be 200-dimensional;
POS, WordNet hyponymy, and grammatical relation embeddings are 50-dimensional.
Each channel of the LSTM network contains the same number of units as its source embeddings
(either 200 or 50). The penultimate hidden layer is 100-dimensional.
As it is not feasible to perform full grid search for all hyperparameters,
the above values are chosen empirically.

We add $\ell_2$ penalty for weights with coefficient $10^{-5}$,
which was chosen by validation
from the set $\{10^{-2}, 10^{-3}, \cdots, 10^{-7}\}$.

We thereafter validate the proposed dropout strategies in Section \ref{ssDropout}.
Since network units in different channels do not interact with each other during information propagation,
we herein take one channel of LSTM networks to assess the efficacy.
Taking the word channel as an example, we first drop out word embeddings.
Then with a fixed dropout rate of word embeddings,  we test the effect of dropping out LSTM inner cells
and the penultimate units, respectively.

We find that, dropout of LSTM units hurts the model,
even if the dropout rate is small, 0.1, say (Figure \ref{fDropout}b).
Dropout of embeddings improves model performance by 2.16\% (Figure \ref{fDropout}a);
dropout of the penultimate layer further improves by 0.16\% (Figure \ref{fDropout}c).
This analysis also provides, for other studies, some clues for dropout in LSTM networks.

\subsection{Results}\label{ssResult}

\begin{table*}[!t]
\centering
\begin{tabular}{c|l|c}
\hline
\hline
\textbf{Classifier} &\centering \textbf{Feature set} &\textbf{$F_1$}\\
\hline
\multirow{3}{*}{SVM}                & POS, WordNet, prefixes and other morphological features, & \multirow{3}{*}{82.2}\\
                   & depdency parse, Levin classes, PropBank, FanmeNet,       &     \\
                   & NomLex-Plus, Google $n$-gram, paraphrases, TextRunner    &     \\
\hline
\multirow{2}{*}{RNN}               & Word embeddings                                          & 74.8\\
                   & Word embeddings, POS, NER, WordNet                       & 77.6\\
\hline
\multirow{2}{*}{MVRNN}              & Word embeddings                                          & 79.1\\
                   & Word embeddings, POS, NER, WordNet                       & 82.4\\
\hline
\multirow{2}{*}{CNN}                &Word embeddings                                           & 69.7\\
                   &Word embeddings, word position embeddings, WordNet        & 82.7\\
\hline
Chain CNN & Word embeddings, POS, NER, WordNet  & 82.7\\
\hline
\multirow{2}{*}{FCM}                & Word embeddings                                         & 80.6\\
                    & Word embeddings, depedency parsing, NER                   & 83.0\\
\hline
\multirow{3}{*}{CR-CNN}              & Word embeddings             & \,\,82.8$^\dag$\\
& Word embeddings, position embeddings & 82.7 \\
                   & Word embeddings, position embeddings                      &\ \ \textbf{84.1}$^\dag$\\
\hline
\multirow{3}{*}{SDP-LSTM}          & Word embeddings                                           & 82.4\\
                    & Word embeddings, POS embeddings, WordNet embeddings,      &
                    \multirow{2}{*}{\textbf{83.7}}\\
                    & grammar relation embeddings                               & \\
\hline
\hline
\end{tabular}
\caption{Comparison of relation classification systems. The ``$\dag$'' remark refers to
special treatment for the {\ttfamily Other} class.}
\end{table*}

Table 4 compares our SDT-LSTM with other state-of-the-art methods.
The first entry in the table presents
the highest performance achieved by traditional feature engineering.
\newcite{2010SVM} leverage a variety of handcrafted features, and use SVM for classification; they achieve an $F_1$-score of 82.2\%.

Neural networks are first used in this task in \newcite{MVRNN}. They build
a recursive neural network (RNN) along a constituency tree for relation classification.
They extend the basic RNN with matrix-vector interaction and achieve an $F_1$-score of 82.4\%.

\newcite{CNN} treat a sentence as sequential data and exploit the convolutional neural network (CNN);
they also integrate word position information into their model. 
\newcite{RankCNN} design a model called CR-CNN;
they propose a ranking-based cost function and
elaborately diminish the impact of the \verb|Other| class, which
is not counted in the official $F_1$-measure.
In this way, they achieve the state-of-the-art result with the $F_1$-score of 84.1\%.
Without such special treatment, their $F_1$-score is 82.7\%.

\newcite{FCM} propose a Feature-rich Compositional Embedding Model (FCM) for relation classification, which combines unlexicalized linguistic contexts and word embeddings. They achieve an $F_1$-score of 83.0\%.

Our proposed SDT-LSTM model yields an $F_1$-score of 83.7\%. It outperforms existing competing
approaches, in a fair condition of softmax with cross-entropy error.

It is worth to note that we have also conducted two controlled experiments: 
(1) Traditional RNN without LSTM units, achieving an $F_1$-score of 82.8\%; 
(2) LSTM network over the entire dependency path (instead of two sub-paths), 
achieving an
$F_1$-score of 82.2\%. These results demonstrate the effectiveness of
LSTM and directionality in 
relation classification.

\subsection{Effect of Different Channels}\label{ssChannel}

This subsection analyzes how different channels affect our model.
We first used word embeddings only as a baseline;
then we added POS tags, grammatical relations, and WordNet hypernyms, respectively;
we also combined all these channels into our models.
Note that we did not try the latter three channels alone, 
because each single of them
(e.g., POS) does not carry much information.


We see from Table \ref{tChannel} that word embeddings alone in SDP-LSTM
yield a remarkable performance of 82.35\%,
compared with CNNs 69.7\%, RNNs 74.9--79.1\%, and FCM 80.6\%.

Adding either grammatical relations or WordNet hypernyms
outperforms other existing methods (data cleaning not considered here).
POS tagging is comparatively less informative, but
still boosts the $F_1$-score by 0.63\%.

We notice that, the boosts are not simply added
when channels are combined. This suggests that these
information sources are complementary to each other in some linguistic aspects.
Nonetheless, incorporating all four channels further
pushes the $F_1$-score to 83.70\%.

\begin{table}[!t]
\centering
\vspace{-.2cm}
\begin{tabular}{lc}
\hline
\hline
\textbf{Channels}  & $F_1$ \\
\hline
Word embeddings     & 82.35  \\
\ $+$ POS embeddings (only)   & 82.98  \\
\ $+$ GR embeddings  (only)   & 83.21   \\
\ $+$ WordNet embeddings (only)    & 83.03\\
\ $+$ POS $+$ GR $+$ WordNet embeddings & 83.70\\
\hline
\hline
\end{tabular}
\caption{Effect of different channels.}\label{tChannel}
\vspace{-.3cm}
\end{table}

\section{Conclusion}\label{sConclusion}
In this paper, we propose a novel neural network model, named SDP-LSTM, for relation classification.
It learns features for relation classification iteratively along the shortest dependency path.
Several types of information (word themselves, POS tags, grammatical relations and WordNet hypernyms) along the path are used.
Meanwhile, we leverage LSTM units for long-range information propagation and integration.
We demonstrate the effectiveness of SDP-LSTM by evaluating the model on SemEval-2010 relation classification task, outperforming 
existing state-of-art methods (in a fair condition without data cleaning).
Our result sheds some light in the relation classification task as follows.

\begin{compactitem}
\item The shortest dependency path can be a valuable resource for relation classification, 
covering mostly sufficient information of target relations.

\item Classifying relation is a challenging task due to the inherent ambiguity of natural languages and the diversity of sentence expression. Thus, 
integrating heterogeneous linguistic knowledge is beneficial to the task.

\item Treating the shortest dependency path as two sub-paths, mapping two different neural networks, helps to capture the directionality of relations.

\item LSTM units are effective in feature detection and propagation along the shortest dependency path.

\end{compactitem}

\section*{Acknowledgments}
This research is supported by the National Basic Research Program of China (the 973 Program) under Grant No. 2015CB352201 and the National Natural Science Foundation of China under Grant Nos. 61232015 and 91318301.

\bibliographystyle{acl}
\bibliography{LSTM}

\end{document}